\documentclass[10pt,twocolumn,letterpaper]{article}

\usepackage{iccv}
\usepackage{times}
\usepackage{graphicx}
\usepackage{amsmath}
\usepackage{amssymb}
\usepackage{booktabs}
\usepackage{xcolor}         
\usepackage{soul}
\usepackage{eqnarray}
\usepackage{algorithm}
\usepackage{algorithmicx}
\usepackage{algpseudocode}
\usepackage{tabularx}
\usepackage{colortbl}
\usepackage{epstopdf}
\usepackage{multirow}
\usepackage{graphicx}
\usepackage{comment}
\usepackage{subfig}
\usepackage{paralist}

\usepackage[pagebackref=true,breaklinks=true,letterpaper=true,colorlinks,bookmarks=false]{hyperref}

\iccvfinalcopy 

\def\iccvPaperID{5334} 

\ificcvfinal\fi

\newcommand{\affmark}[1][*]{\textsuperscript{#1}}

\begin{document}

\title{\vspace{-25pt}ALUM: Adversarial Data Uncertainty Modeling from Latent Model Uncertainty Compensation}

\author{Wei Wei\affmark[1], Jiahuan Zhou\affmark[2]\thanks{Jiahuan Zhou is the corresponding author.}, Hongze Li\affmark[2], and Ying Wu\affmark[1] \\
\affmark[1]Department of Electrical and Computer Engineering, Northwestern University, Evanston, IL, USA\\
\affmark[2]Wangxuan Institute of Computer Technology, Peking University, Beijing, China\\
{\tt\small wwzjer@u.northwestern.edu, jiahuanzhou@pku.edu.cn,  lihongze@pku.edu.cn, yingwu@northwestern.edu}}

\maketitle

\begin{abstract}
    It is critical that the models pay attention not only to accuracy but also to the certainty of prediction. Uncertain predictions of deep models caused by noisy data raise significant concerns in trustworthy AI areas. To explore and handle uncertainty due to intrinsic data noise, we propose a novel method called ALUM to simultaneously handle the model uncertainty and data uncertainty in a unified scheme. Rather than solely modeling data uncertainty in the ultimate layer of a deep model based on randomly selected training data, we propose to explore mined adversarial triplets to facilitate data uncertainty modeling and non-parametric uncertainty estimations to compensate for the insufficiently trained latent model layers. Thus, the critical data uncertainty and model uncertainty caused by noisy data can be readily quantified for improving model robustness. Our proposed ALUM is model-agnostic which can be easily implemented into any existing deep model with little extra computation overhead. Extensive experiments on various noisy learning tasks validate the superior robustness and generalization ability of our method. The code is released at \url{https://github.com/wwzjer/ALUM}.
\end{abstract}

\section{Introduction}

\begin{figure*}[t]
\vspace{-20pt}
\begin{center}
\includegraphics[width=\textwidth,height=160pt]{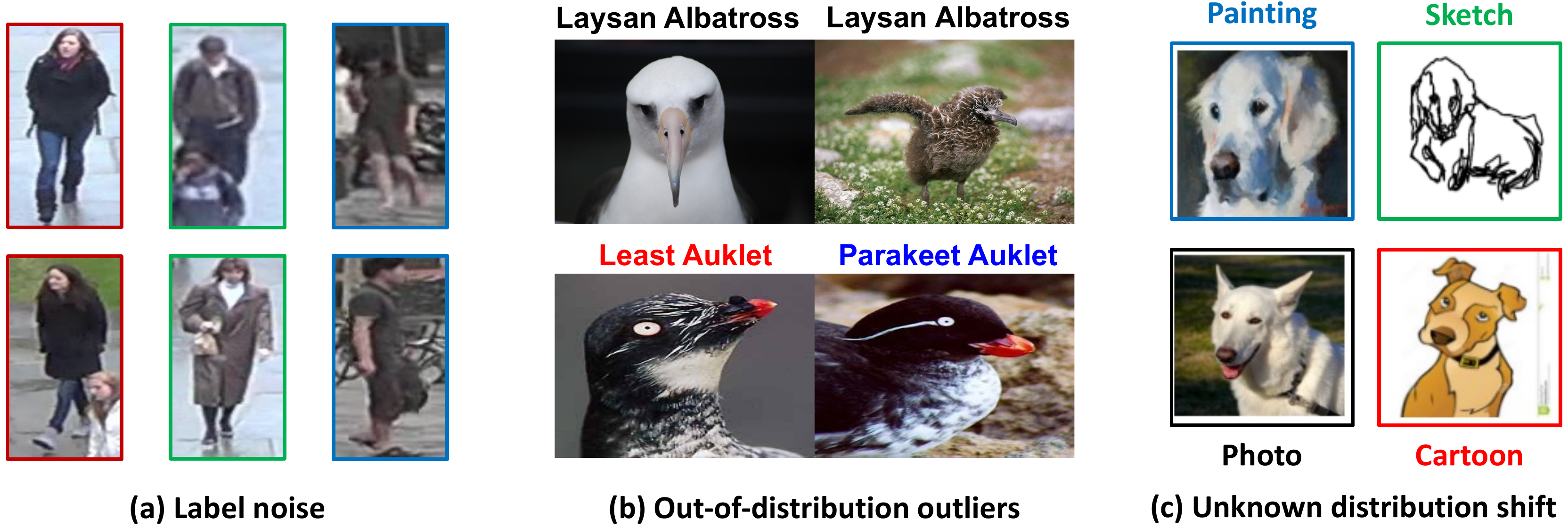}
\end{center}
\vspace{-15pt}
\caption{\label{fig:noise}In the real world, the collected training data always suffer from various data noise caused by (a) \textbf{Label noise}: different people in the first and second rows are incorrectly labeled as having the same identity because of similar clothing, (b) \textbf{Out-of-distribution outliers}: the same species of birds may have large differences in morphology, while different species may be very similar, and (c) \textbf{Unknown distributions shifting}: training and testing data share distinct domains.}
\vspace{-15pt}
\end{figure*}

In real life, \textit{the only certainty is uncertainty}. In machine learning, uncertainty is generally split into model uncertainty (\emph{epistemic}) and data uncertainty (\emph{aleatoric})~\cite{der2009aleatory,hullermeier2021aleatoric}. More specifically, model uncertainty is usually due to insufficient parameter training caused by imperfect training data (\eg, small-size data, degraded data, noisy data, \etc). Thus, a general solution to mitigate model uncertainty is utilizing more training data, but it will introduce a heavier computational burden as well. On the other hand, data uncertainty mainly comes from intrinsic data noise caused by various real-world factors, such as label noise (Fig.~\ref{fig:noise}(a)), out-of-distribution outliers (Fig.~\ref{fig:noise}(b)), and unknown distribution shift (Fig.~\ref{fig:noise}(c)). Therefore, learning deep models from noisy data always leads to unreliable predictions that inevitably limit their robustness capability and may result in disastrous consequences in many trustworthy AI tasks, such as autonomous driving~\cite{feng2018towards,choi2019gaussian,cai2021yolov4}, medical diagnosis~\cite{seebock2019exploiting,zhang2019reducing,nair2020exploring}, and security identification~\cite{yu2019robust,zhou2017efficient,zou2019learning}.

As an important and long-standing issue, how to improve model robustness and suppress the adverse effects of uncertainty has attracted considerable attention. In recent years, many efforts have been dedicated to the exploration of \textbf{uncertainty modeling}~\cite{ghahramani2015probabilistic,blundell2015weight,gal2016dropout,lakshminarayanan2017simple,malinin2018predictive,oh2018modeling,yu2019robust,van2020uncertainty,zhang2021relative,li2022uncertainty}. Instead of representing training data as deterministic feature vectors, uncertainty modeling proposes to learn a probabilistic multivariate Gaussian distribution for each datum with the mean depicting the spatial location in the learned feature space and the variance measuring uncertainty. To do so, two separate streams representing the mean and variance are usually concatenated after the penultimate layer of a deep model as the uncertainty learning branch. However, recent work~\cite{zhang2021relative} has verified that the superior learning ability of deep models will deteriorate the performance of uncertainty learning, evidenced by the phenomenon that the variance vector will rapidly converge to zero so that it will be no longer beneficial to the remaining learning epochs. Although several latest studies~\cite{zhang2021relative,she2021dive} have tried to utilize pair-wise uncertainty normalization to mitigate the variance diminishing issue, they still suffer from deteriorated uncertainty learning performance, due to the randomness of their pair-wise data selection mechanism, without thorough exploitation of information among training data. 

In this paper, we accordingly propose a novel method via leveraging the informative merits of \textbf{A}dversarial examples in training data and \textbf{L}atent features of intermediate layers in deep models to facilitate \textbf{U}ncertainty \textbf{M}odeling, named \textbf{ALUM} for abbreviation. With the assumption that the small inter-class variations and large intra-class variations lead to uncertainty in learning~\cite{zhang2019mitigating}, our ALUM approach aims to find and leverage these ``adversarial examples'' from given training data to reduce the uncertainty. Compared with existing adversarial training methods~\cite{Goodfellow2015Explaining,madry2018towards,zhang2019theoretically} which are severely time-consuming, we directly mine two adversarial batches from the given training batch itself as a triplet which is efficient and effective. In addition, ALUM simultaneously regularizes the model uncertainty of the learned deep model by using non-parametric uncertainty estimation as compensation for all the channels in an intermediate layer. In this way, the adverse effects of model uncertainty will be appropriately suppressed during training, facilitating the probabilistic data uncertainty modeling. To sum up, our main contributions are three-fold:

1. To alleviate the adverse effect of data uncertainty caused by noisy data, we propose ALUM to mine and leverage adversarial examples from training data to adapt to the superior learning capability of deep models as well as to avoid deteriorating the performance of uncertainty learning. 

2. In addition to modeling data uncertainty in the ultimate feature embedding space, our ALUM also simultaneously quantifies the model uncertainty of intermediate network layers by compensating the latent feature maps based on non-parametric uncertainty estimation.

3. Extensive experiments on various noisy learning tasks validate the effectiveness of our method. Compared with state-of-the-art uncertainty modeling methods, ALUM exhibits enhanced robustness to various data noise conditions.

\section{Related Work}

\noindent
\textbf{Uncertainty Modeling for Learning from Noise.} Data noise naturally exists in many computer vision and machine learning tasks, thus various uncertainty modeling methods are proposed to tackle this critical issue. We focus on \textit{classification and recognition} tasks in this paper. \cite{oh2018modeling} utilized the variational information bottleneck principle to model a category-level uncertainty to facilitate classification. 
Following a recent multi-domain generalization method pAdaIN~\cite{nuriel2021permuted}, \cite{li2022uncertainty} modeled the uncertainty of domain shifts using synthesized feature statistics during training to address the multi-domain classification problem. 
\cite{yang2022domain} utilized style noise from various domains to augment the input, for the purpose of regularizing the encoder to gain domain-invariant features. \cite{meng2022attention} proposed a paradigm to simulate domain shift by dividing attention maps into task-related and domain-related groups.
To mitigate the influence of noisy data in facial expression recognition, \cite{wang2020suppressing} designed a self-cure network to suppress the uncertainties of facial expression data and thus learn robust features. Besides, \cite{zhang2022learn} designed an imbalanced framework about flip semantic consistency and randomly erase noisy images to motivate the model to focus on global features. Along with \cite{wang2020suppressing}, \cite{zhang2021relative} further proposed an innovative uncertainty learning method focusing on modeling the relative uncertainty between pairwise samples to reduce the impact of noise.

Nevertheless, these uncertainty modeling methods only focus on modeling and quantifying either model uncertainty or data uncertainty in the ultimate feature space. Our proposed ALUM can simultaneously handle model uncertainty via a latent non-parametric feature compensation and tackle data uncertainty via adversarial triplet-driven uncertainty learning. Therefore, our ALUM not only fully explores the helpful adversarial examples in the given training data but also takes the informative latent features in the intermediate layers into consideration. As a result, the robustness and discrimination of the learned deep model can be improved.

\noindent
\textbf{Adversarial Learning and Feature Mixup.} Besides uncertainty modeling, many strategies have been developed to improve deep model robustness. Among them, \textbf{adversarial learning} and \textbf{feature mixup} are widely explored. The core purpose of adversarial learning is to make networks robust toward adversarial examples which are visually similar to natural input but with small imperceptible perturbation. Since raised by~\cite{biggio2013evasion,Szegedy2014Intriguing,Goodfellow2015Explaining}, adversarial training is regarded as one of the most effective methods for model robustness enhancement. The promising performance is mainly caused by generating sufficient adversarial examples during training to substitute or combine with natural clean data for model training~\cite{madry2018towards,zhang2019theoretically}. However, adversarial example generation processing in most adversarial training methods severely suffers from high computational cost and extra data burden, while our ALUM efficiently and effectively mines the adversarial examples from the given training dataset itself, framed under the scope of uncertainty modeling.

As a useful data augmentation strategy for learning, Mixup~\cite{zhang2018mixup} aimed to produce the paired samples which is the convex combination of two independent samples and corresponding labels. Later, different variants of mixup including Manifold Mixup \cite{verma2019manifold}, Cutmix \cite{yun2019cutmix}, and Noisy Feature Mixup \cite{lim2022noisy} were proposed to further gain better performance. In our proposed ALUM, we expect to borrow the data augmentation merits of the mixup. While significantly different from the above mixup methods, our mixup coefficients are not rooted in randomized weighting parameters but are obtained from the uncertainty modeling results. Therefore, the uncertainty predictions of training data can be thoroughly explored to benefit the uncertainty learning performance further.

\begin{figure*}
\vspace{-10pt}
    \begin{center}
	\includegraphics[width=1\textwidth]{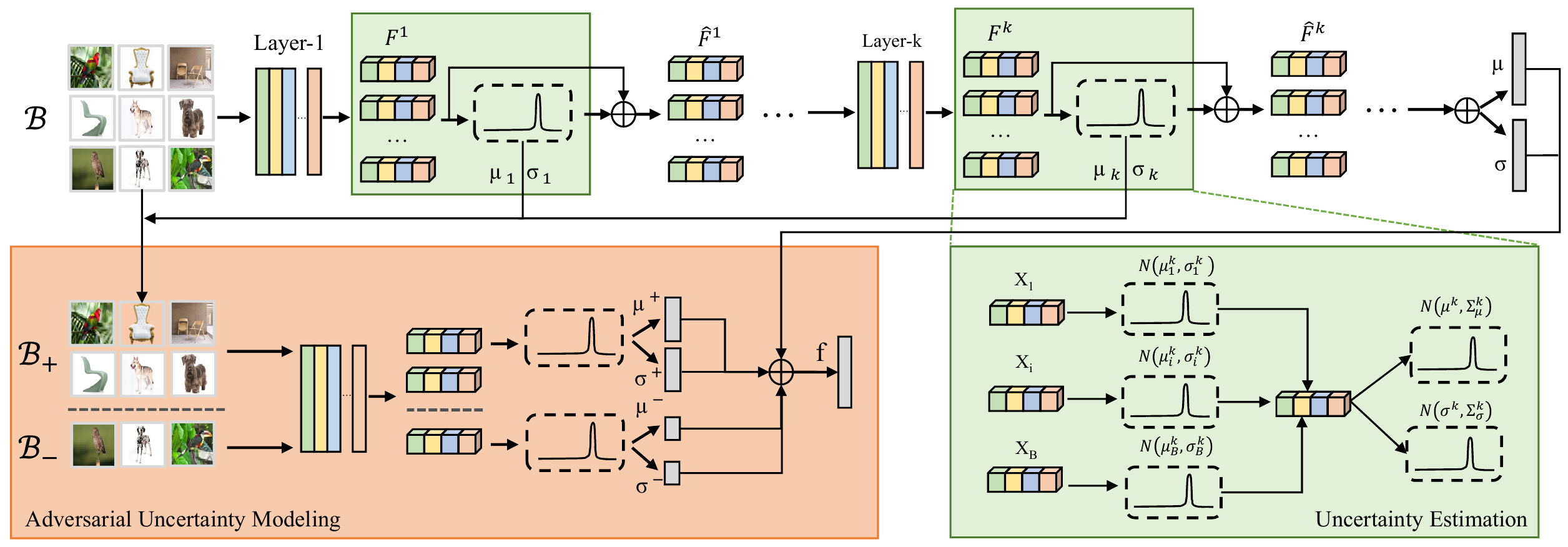}
        \vspace{-10pt}
        \caption{\label{fig:framework} Our proposed ALUM network models adversarial data uncertainty based on latent model uncertainty compensations. For each intermediate layer in the network, its non-parametric channel-wise uncertainty estimation is obtained as the compensation to refine the extract latent feature maps (with dimension $C \times H \times W$) for the fed training samples (with batch size $B$).}
        \vspace{-15pt}
    \end{center}
\end{figure*}

\section{The Proposed ALUM Method}

In this section, we demonstrate the overall structure of our proposed ALUM method in detail, as shown in Fig.~\ref{fig:framework}. Firstly, a channel-wise non-parametric uncertainty estimation module (Sec.~\ref{sec:channel}) is utilized to characterize the model uncertainty of each intermediate layer. Then, based on the above channel-wise uncertainty estimation, a layer-wise uncertainty-aware transformation (Sec.~\ref{sec:layer}) is adopted to compensate for the intermediate learning results of each layer. In the ultimate layer, two batches of adversarial positives and negatives are mined based on the compensated feature maps to intrinsically form a triplet for the final parametric adversarial uncertainty modeling (Sec.~\ref{sec:final}).

\subsection{Channel-Wise Non-Parametric Uncertainty Estimation}
\label{sec:channel}
A batch of training data is denoted as $\mathcal{B} = \{(x_i,l_i)\}_{i=1}^B$ where $l_i$ indicates the class label of $x_i$, but may exhibit label noise. Then $\mathcal{B}$ is fed into a deep model $\mathcal{M}_{\theta}$ whose feature outputs of the $k$-th intermediate layer $\mathcal{M}^k_\theta$ can be represented as:

\begin{equation}
    \label{eqn:latent-maps}
   \mathcal{F}^k = \mathcal{M}_{\theta}^k(\mathcal{F}^{k-1}), ~~\mathcal{F}^k \in \mathbb{R}^{B \times C \times H \times W},
\end{equation}
where the feature extraction of $\mathcal{B}$ from the $k$-th intermediate layer consists of $C$ channels that each channel is a $H \times W$ feature map. Furthermore, motivated by \cite{nuriel2021permuted,li2022uncertainty}, we suppose that the uncertainty of each channel follows a Gaussian distribution. 
So the channel-wise non-parametric spatial feature uncertainty estimation $\mathcal{N}(\mu^{ck},\sigma^{ck})$ for the $b$-th sample's $c$-th channel of the $k$-th intermediate layer can be calculated as:

{\small
\begin{align}
    \label{eqn:channel}
    \mu^{ck} &= \dfrac{1}{HW}\sum_{h=1}^H\sum_{w=1}^W\mathcal{F}^k_{b,c,h,w}, \\
    \sigma^{ck} &= \sqrt{\dfrac{1}{HW}\sum_{h=1}^H\sum_{w=1}^W \left( \mathcal{F}^k_{b,c,h,w} - \mu^{ck} \right)^2}.
\end{align}
}

By performing Eqn.~\ref{eqn:channel} over all the $B$ samples in $\mathcal{B}$ for all the $C$ channels of the $k$-th intermediate layer, we could obtain the uncertainty estimation results with the mean $\mathcal{U}^k$ and the variance $\mathcal{S}^k$:

\begin{equation}
    \mathcal{U}^k \in \mathbb{R}^{B \times C}, ~~ \mathcal{S}^k \in \mathbb{R}^{B \times C}.
\end{equation}

As shown in Fig.~\ref{fig:framework}, we further perform a similar channel-wise non-parametric uncertainty estimation for $\mathcal{U}^k$ and $\mathcal{S}^k$ over all batch samples which are denoted as $\mathcal{N}(\mu^k, \Sigma_\mu^k)$ and $\mathcal{N}(\sigma^k, \Sigma_\sigma^k)$ in Eqn.~\ref{eqn:mu} and Eqn.~\ref{eqn:sigma} respectively:

\begin{equation}
\label{eqn:mu}
    \mu^k = \dfrac{1}{B}\sum_{b=1}^B\mathcal{U}^k_b, ~~ \Sigma_\mu^k = \sqrt{\dfrac{1}{B}\sum\limits_{b=1}^B \left(\mathcal{U}^k_b - \mu^k\right)^2},
\end{equation}
\begin{equation}
\label{eqn:sigma}
    \sigma^k = \dfrac{1}{B}\sum_{b=1}^B\mathcal{S}^k_b, ~~ \Sigma_\sigma^k = \sqrt{\dfrac{1}{B}\sum\limits_{b=1}^B \left(\mathcal{S}^k_b - \sigma^k\right)^2}.
\end{equation}

\subsection{Layer-Wise Uncertainty-Aware Compensation}
\label{sec:layer}
The uncertainty estimation results $\mathcal{N}(\mu^k, \Sigma_\mu^k)$ and $\mathcal{N}(\sigma^k, \Sigma_\sigma^k)$ play a crucial role to characterize the statistical properties of the latent feature spaces. We aim to utilize them to compensate for the model uncertainty caused by noisy data to further improve the generalization and robustness of the learned deep model. Specifically, we explore $\mathcal{N}(\mu^k, \Sigma_\mu^k)$ and $\mathcal{N}(\sigma^k, \Sigma_\sigma^k)$ in \cite{li2022uncertainty} to perform a layer-wise uncertainty-aware transformation for the latent feature maps $\mathcal{F}^{k}$ as below:

\begin{equation}
    \label{eqn:compensatrion}
    \hat{\mathcal{F}}^{k} = \left( \sigma^{k} + \epsilon\Sigma_\sigma^k \right) \left( \dfrac{\mathcal{F}^{k} - \mu^{k}}{\sigma^{k}} \right)  + \left( \mu^{k} + \epsilon\Sigma_\mu^k\right).
\end{equation}

The reformed intermediate feature outputs $\hat{\mathcal{F}}^k$ is compensated by a random perturbation $\epsilon$ of $\Sigma_\sigma^k$ and $\Sigma_\mu^k$, then it will be further forwarded to the next layers:

\begin{equation}
    \label{eqn:next}
    \mathcal{F}^{k+1} = \mathcal{M}_{\theta}^{k+1}(\hat{\mathcal{F}}^k).
\end{equation}

\subsection{Parametric Adversarial Uncertainty Modeling}
\label{sec:final}
Based on Eqn.~\ref{eqn:next}, once the training batch $\mathcal{B}$ is fed until the penultimate layer $L-1$ to obtain feature maps $\mathcal{F}^{L-1}$, a two-branch architecture is concatenated on the top of $\mathcal{F}^{L-1}$ to predict a mean vector $\mu_i$ and a variance vector $\sigma_i$ for $x_i$ respectively:

\begin{equation}
    \label{eqn:us}
    \mathcal{U} = \{\mu_i\}_{i=1}^B, ~~ \mathcal{S} = \{\sigma_i\}_{i=1}^B.
\end{equation}

Instead of directly using $\mathcal{U}$ as the feature representation for training, we further propose an adversarial example mining-based uncertainty-aware feature mixup model to facilitate the learning of $\mathcal{B}$. Specifically, two adversarial counterpart batches, a positive one $\mathcal{B}_{+} = \{(x_i^+, l_i^+)\}_{i=1}^B$ and a negative one $\mathcal{B}_{-} = \{(x_i^-, l_i^-)\}_{i=1}^B$, are used to form a triplet $\mathcal{T}_i = (x_i, x_i^+, x_i^-)$ for each $x_i$ by solving Eqn.~\ref{eqn:adversarial}:

\begin{equation}
\label{eqn:adversarial}
    x_i^+ = \underset{j,~ \text{s.t.}~ l_j=l_i}{\mathrm{argmax}}~D(\mu_i,\mu_j), ~~ x_i^- = \underset{j, ~ \text{s.t.}~ l_j\ne l_i}{\mathrm{argmin}}~D(\mu_i,\mu_j),
\end{equation}
where $D$ is the cosine similarity distance and $j \ne i$. Thus, $\mathcal{B}_{+}$ contains the adversarial positives which are visually distinct from $\mathcal{B}$, and $\mathcal{B}_{-}$ contains the adversarial negatives which are visually similar to $\mathcal{B}$. On one hand, $\mathcal{B}_{+}$ and $\mathcal{B}_{-}$ are informative enough to mitigate the variance diminishing problem. On the other hand, these adversarial examples inevitably deteriorate the convergence of model training since they are very hard to handle. Therefore, inspired by the self-paced learning theory~\cite{kumar2010self}, we adopt the adversarial examples $\mathcal{B}_{+}$ and $\mathcal{B}_{-}$ in a ``from-easy-to-hard'' manner. A parameter $p$ is used to control the percentage of how many adversarial samples are mined, and the other samples are the randomly shuffled results of $\mathcal{B}$.

\begin{algorithm}[!tbp]
	\caption{\label{alg:ALUM} Algorithm of the Proposed ALUM Method}
	\begin{algorithmic}[1]
		\Require A training batch $\mathcal{B} = \{(x_i,l_i)\}_{i=1}^B$.
		\Ensure The well-trained uncertainty modeling network $\mathcal{M}_{\theta}$ with $L$ layers.
		\State $\mathcal{B}$ is fed into $\mathcal{M}_{\theta}(\mathcal{B})$ and obtain the feature maps from the first layer $\mathcal{F}^1 = \mathcal{M}_{\theta}^1(\mathcal{B})$;
		\For{each intermediate layers $k \le L-1$ of $\mathcal{M}_{\theta}$}
		\State Compute the latent feature maps $\mathcal{F}^k$ from $k$-th layer via Eqn.~\ref{eqn:latent-maps};
		\State Compute the channel-wise non-parametric uncertainty estimation $\mathcal{U}^k$ and $\mathcal{S}^k$ via Eqn.~\ref{eqn:channel};
		\State Obtain the compensated feature maps $\hat{\mathcal{F}}^{k}$ in $k$-th layer via Eqn.~\ref{eqn:compensatrion};
		\State Feed $\hat{\mathcal{F}}^{k}$ to the next layer via Eqn.~\ref{eqn:next};
		\EndFor
		\State In the ultimate $L$-th layer, obtain the initial uncertainty modeling results $\mathcal{U}$ and $\mathcal{S}$ of $\mathcal{B}$ as Eqn.~\ref{eqn:us};
		\State Mine one adversarial positive batch $\mathcal{B}_{+}$ and one adversarial negative batch $\mathcal{B}_{-}$ via Eqn~\ref{eqn:adversarial};
		\State Perform Hadamard operation-based feature mixup via Eqn.~\ref{eqn:f_i};
		\State Optimize the overall learning loss $\mathcal{L}$ in Eqn.~\ref{eqn:loss} until convergence.
	\end{algorithmic}
\end{algorithm}

Hence, denote the probabilistic uncertainty predictions of $x_i$, $x_i^+$, and $x_i^-$  as $\mathcal{N}(\mu_i, \sigma_i)$, $\mathcal{N}(\mu_i^+, \sigma_i^+)$ and $\mathcal{N}(\mu_i^-, \sigma_i^-)$ respectively, an uncertainty-aware feature mixup is performed to $\mathcal{U}$ via a Hadamard operation-based normalization processing:

\begin{equation}
\label{eqn:f_i}
    f_i = w_i\circ\mu_i + w_i^+\circ\mu_i^+ + w_i^-\circ\mu_i^-,
\end{equation}
where $w_i = \sigma_i \oslash (\sigma_i+\sigma_i^++\sigma_i^-)$, so as to $w_i^+$ and $w_i^-$. Noted that $\circ$ and $\oslash$ represent the element-wise Hadamard multiplication and division respectively. The mixed feature $f_i$ should be discriminative enough to recognize $l_i$, $l_i^+$, and $l_i^-$ classes individually and jointly. To do so, a learning loss $\mathcal{L}$ in Eqn.~\ref{eqn:loss} consisting of a cross-entropy loss (Eqn.~\ref{eqn:loss_s}) and a triplet loss (Eqn.~\ref{eqn:loss_t}) is optimized:

\begin{equation}
    \label{eqn:loss}
    \mathcal{L} = \mathcal{L}_C + \lambda\mathcal{L}_T,
\end{equation}
and the two loss terms are formulated as below:
{\scriptsize
\begin{equation}
\label{eqn:loss_s}
   \mathcal{L}_C =  - \dfrac{1}{B}\sum\limits_{i=1}^B {\left( {\log \dfrac{{{e^{{W_i}f_i}}}}{{\sum\limits_c {{e^{{W_c}f_i}}} }} + \log \dfrac{{{e^{{W_i^+}f_i}}}}{{\sum\limits_c {{e^{{W_c}f_i}}} }} + \log \dfrac{{{e^{{W_i^-}f_i}}}}{{\sum\limits_c {{e^{{W_c}f_i}}} }}} \right)},
\end{equation}
}
\begin{equation}
\label{eqn:loss_t}
    \mathcal{L}_T = \sum\limits_{i=1}^B \max \left(\|\mu_i - \mu_i^+\|_2^2 - \|\mu_i - \mu_i^-\|_2^2 +\alpha, 0 \right),
\end{equation}
where ${W_i}$ is the classifier for the $i$-th class, and $\alpha$ is the margin parameter set to 1 by default. The cross-entropy loss is to recognize the triplet inputs jointly from the mixed features, while the triplet loss is to enforce the compactness between samples with the same label and the separation between samples with different labels. The overall algorithm of our proposed ALUM method is summarized in Alg.~\ref{alg:ALUM}.

\section{Experiments}
\label{sec:exp}

In this section, to verify the effectiveness of our proposed ALUM method, we conduct extensive experiments on a wide range of evaluation tasks including facial expression recognition under label noise (Sec.~\ref{sec:exp-fer}), cross-domain generalization (Sec.~\ref{sec:exp-multidomain}), general image classification (Sec.~\ref{sec:exp-general}) and more ablation study experiments (Sec.~\ref{sec:exp-ablation}).

\subsection{Noisy Facial Expression Recognition}
\label{sec:exp-fer}

\noindent
\textbf{Datasets}: Facial expression recognition (FER) is a typical task that suffers from label noise due to annotator subjectivity and uncertainty introduced by in-the-wild face images. To verify the effectiveness of our proposed ALUM on handling \textbf{label noise}, we conducted experiments on one of the most widely-used FER datasets: Real-world Affective Faces Database (RAF-DB~\cite{li2017reliable}, which contains 29672 facial images annotated with basic and compound expressions. In our experiments, we adopt 7 basic expressions (\ie neutral, happiness, surprise, sadness, anger, disgust, and fear) including 12271 training images and 3068 test images. 

\noindent
\textbf{Settings}: Following~\cite{zhang2021relative}, all images in RAF-DB are resized to 224$\times$224, and the ResNet-18~\cite{he2016deep} pretrained on Ms-Celeb-1M~\cite{guo2016ms} is used as the backbone network. To train our ALUM model, the training and testing batch size is 128, the total number of training epochs is 60, and the learning rate is initialized as 0.0004 and 0.004 for the backbone layers and the last fully connected layer respectively. Besides, an Adam optimizer~\cite{kingma2014adam} with a weight decay of 0.0001 is used. The weighting parameter $\lambda$ is 0.003 for balancing loss scale, adversarial positive/negative controlling parameter $p$ is 0.2. All the experiments are run on a single NVIDIA GeForce 3090 GPU with 24GB memory.

\noindent
\textbf{Evaluation on Learning from Noisy Labels}: In Table~\ref{tab:raf-db}, several state-of-the-art noise-tolerant facial expression recognition methods are compared in the experiments including SCN~\cite{wang2020suppressing}, DUL~\cite{chang2020data}, DMUE~\cite{she2021dive}, and RUL~\cite{zhang2021relative}. Besides the ambiguous outlier images in training data (Noisy 0\%), we follow \cite{chang2020data,wang2020suppressing,zhang2021relative} to randomly choose 10\%, 20\%, 30\% of training data and randomly flip their labels to other categories, which can generate inconsistent label noise for learning.  For a fair comparison, the same backbone of ResNet-18 and pretrained model are utilized. For the latest RUL method, which can be seen as our baseline, we exactly follow their released code with the optimal training settings and parameters reported in~\cite{zhang2021relative} to reproduce their results (the reported results in~\cite{zhang2021relative} are also included). As reported in Table~\ref{tab:raf-db}, our ALUM significantly outperforms the other methods under all circumstances by consistently improving recognition accuracy with different ratios of noisy labels. Notably, the improvement on the 30\% is most significant, with over 4\%. Thus our ALUM can benefit more as the noise ratio increases which demonstrates that ALUM is more robust to noisy labels. In comparison with our baseline RUL, it verifies the effectiveness of our adversarial data uncertainty modeling and latent model uncertainty compensation.

Moreover, to benchmark the performance on RAF-DB, we compare our method with the recent SOTA method EAC~\cite{zhang2022learn}. In Table~\ref{tab:raf-db sota}, the same backbone of ResNet-50 is adopted for EAC and ALUM. Again, our ALUM achieves better performance on different noisy levels.

\begin{table}[t]
\begin{center}
\vspace{-10pt}
{\small
\begin{tabular}{l|c|c|c}
\toprule
    Method & Reference & Noisy(\%) & Accuracy(\%)\\
    \midrule
    SCN~\cite{wang2020suppressing} & CVPR’2020 & 0 & 87.35 \\
    DUL~\cite{chang2020data}       & CVPR’2020 & 0 & 88.04 \\
    DMUE~\cite{she2021dive}        & CVPR’2021 & 0 & 88.76 \\
    RUL$\star$~\cite{zhang2021relative}   & NeurIPS’2021 & 0 & 88.98 \\
    RUL$\dagger$~\cite{zhang2021relative}   & NeurIPS’2021 & 0 & 88.33 \\
    \midrule
    \textbf{ALUM}   & \textbf{Ours}& 0 & \textbf{89.90} \\
\toprule
    SCN~\cite{wang2020suppressing} & CVPR’2020 & 10 & 81.92 \\
    DUL~\cite{chang2020data}       & CVPR’2020 & 10 & 85.08 \\
    DMUE~\cite{she2021dive}        & CVPR’2021 & 10 & 83.19 \\
    RUL{$\star$}~\cite{zhang2021relative}   & NeurIPS’2021 & 10 & 86.22 \\
    RUL{$\dagger$}~\cite{zhang2021relative}   & NeurIPS’2021 & 10 & 87.13 \\
    \midrule
    \textbf{ALUM}   & \textbf{Ours} & 10 & \textbf{87.68} \\
\toprule
    SCN~\cite{wang2020suppressing}  & CVPR’2020 & 20 & 80.02 \\
    DUL~\cite{chang2020data}        & CVPR’2020 & 20 & 81.95 \\
    DMUE~\cite{she2021dive}         & CVPR’2021 & 20 & 81.02 \\
    RUL{$\star$}~\cite{zhang2021relative}    & NeurIPS’2021 & 20 & 84.34 \\
    RUL{$\dagger$}~\cite{zhang2021relative}    & NeurIPS’2021 & 20 & 84.94 \\
    \midrule
    \textbf{ALUM}   & \textbf{Ours} & 20 & \textbf{85.37} \\
\toprule
    SCN~\cite{wang2020suppressing}  & CVPR’2020 & 30 & 77.46 \\
    DUL~\cite{chang2020data}        & CVPR’2020 & 30 & 78.90 \\
    DMUE~\cite{she2021dive}         & CVPR’2021 & 30 & 79.41 \\
    RUL{$\star$}~\cite{zhang2021relative}    & NeurIPS’2021 & 30 & 82.06 \\
    RUL{$\dagger$}~\cite{zhang2021relative}    & NeurIPS’2021 & 30 & 80.87 \\
    \midrule
    \textbf{ALUM}   & \textbf{Ours} & 30 & \textbf{84.91} \\
\bottomrule
\end{tabular}}
\vspace{-5pt}
\caption{\label{tab:raf-db}Experiment results on facial expression datasets RAF-DB with backbone of ResNet-18. ($\star$: reported results; $\dagger$:  reproduced results exactly following the released code.)}
\end{center}
\vspace{-10pt}
\end{table}

\begin{table}[t]
\begin{center}
\vspace{-10pt}
{\small
\begin{tabular}{l|c|c|c}
\toprule
    Method & Reference & Noisy(\%) & Accuracy(\%) \\
    \midrule
    EAC~\cite{zhang2022learn}  & ECCV'2022 & 0 & 90.35\\
    \textbf{ALUM}   & \textbf{Ours} & 0 & \textbf{90.51}  \\
\toprule
    EAC~\cite{zhang2022learn}  & ECCV'2022 & 10 & 88.62 \\
    \textbf{ALUM}   & \textbf{Ours} & 10 & \textbf{89.31}  \\
\toprule
    EAC~\cite{zhang2022learn}  & ECCV'2022 & 20 & 87.35 \\
    \textbf{ALUM}   & \textbf{Ours} & 20 & \textbf{88.07} \\
\toprule
    EAC~\cite{zhang2022learn}  & ECCV'2022 & 30 & 85.27 \\
    \textbf{ALUM}   & \textbf{Ours} & 30 & \textbf{86.34} \\
\bottomrule
\end{tabular}}
\vspace{-5pt}
\caption{\label{tab:raf-db sota}Benchmarking on RAF-DB over the state-of-the-art method, with the backbone of ResNet-50.}
\end{center}
\vspace{-20pt}
\end{table}

\noindent
\textbf{Evaluation on Accuracy-Rejection Tradeoff}: To further verify that ALUM can predict meaningful uncertainty values and utilize them to identify the most uncertain data, we conduct experiments using the \textit{accuracy versus rejection rate}~\cite{zhang2021relative} metric: according to the uncertainty predictions, the images with top 10\%, 20\%, and 30\% uncertain values are directly rejected, and the test accuracy is calculated on the remaining images. Therefore, the better a method models the uncertainty, the better test accuracy can be achieved consistently when increasing the rejection ratio. As reported in Table~\ref{tab:rejection}, our ALUM outperforms all the other methods in all settings. This result strongly demonstrates our ALUM can effectively learn more meaningful uncertainty values which are closely and consistently related to the final recognition confidence.

\begin{figure*}[t]
\vspace{-15pt}
\begin{center}
    \includegraphics[width=\textwidth,height=90pt]{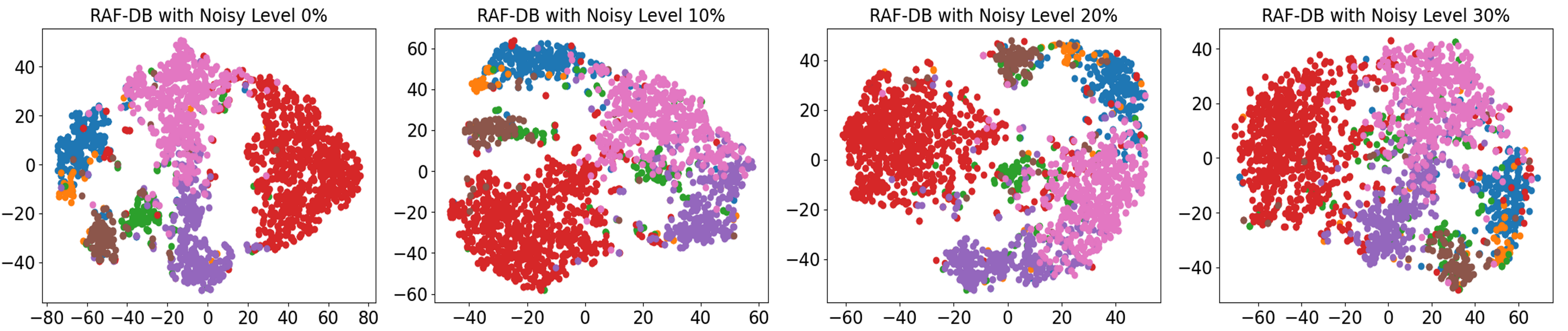}
\end{center}
\vspace{-15pt}
\caption{\label{fig:tsne}The t-SNE results of test feature embedding learned by ALUM on RAF-DB with noise level of 0-30\%.}
\vspace{-10pt}
\end{figure*}

\noindent
\textbf{The t-SNE Visualization of Learned Features}: To further verify our proposed ALUM is robust to data noise, we utilize t-SNE~\cite{van2008visualizing} to visualize the learned features of our ALUM under different noise levels on RAF-DB. As shown in Fig.~\ref{fig:tsne}, the feature embedding space learned by our ALUM demonstrates the intra-category compactness and inter-category separability even if the noise level is at a high-level 30\%. This visualization result is consistent with previous quantitative results which further demonstrates the proposed ALUM is robust to data noise.

\begin{figure*}[htbp]
  \begin{center}
      \includegraphics[width=0.95\linewidth]{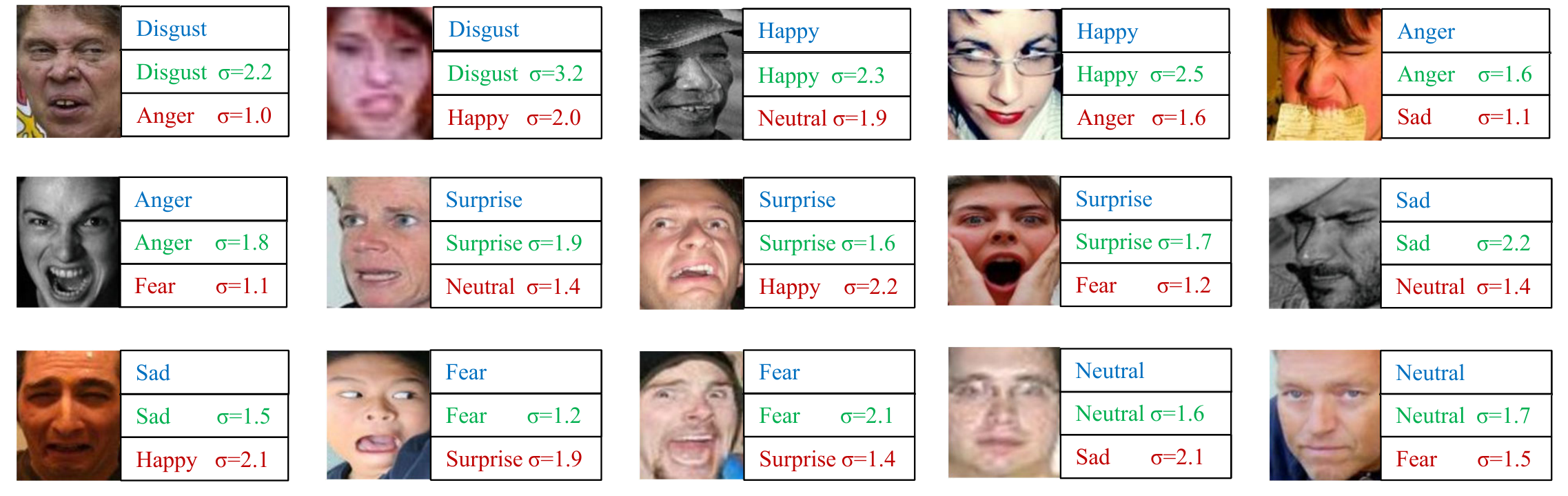}
  \end{center}
  \vspace{-15pt}
  \caption{\label{fig:visual}The visualization results of our proposed ALUM and the baseline method RUL on some hard examples in RAF-DB.}
  \vspace{-10pt}
\end{figure*}

\noindent
\textbf{The Visualization of Learned Uncertainty Values}:
To further verify that our proposed ALUM can better quantify the uncertainty of these error-prone samples, we conducted the following visualization experiments. Fig.~\ref{fig:visual} compares the uncertainty scores and recognition results of our ALUM and baseline method RUL~\cite{zhang2021relative} for the same case. The ground-truth label of each case is shown in \textcolor{blue}{Blue} color, and the prediction results of ALUM are marked in \textcolor{green}{Green} and RUL in \textcolor{red}{Red} respectively. For some examples, although RUL predicts relatively small uncertainty values for them by considering them as easy cases, RUL still can not correctly classify them which demonstrates the uncertainty estimation results of RUL deviate from the factual results. While our ALUM performs well on the examples with either high uncertainty scores or low scores. This demonstrates that our ALUM can not only accurately estimate the uncertainty of the images, but also correctly classify them by eliminating the adverse influence of uncertainty.

\begin{table}[t]
\vspace{-10pt}
\begin{center}
{\small
\begin{tabular}{l|cccc}
\toprule
Rejection & 0\% & 10\% & 20\% & 30\% \\
\midrule
SCN~\cite{wang2020suppressing}   & 87.35 & 86.85 & 86.63 & 87.28 \\
DUL~\cite{chang2020data}         & 88.04 & 90.11 & 92.58 & 94.50 \\
RUL$\dagger$~\cite{zhang2021relative}     & 88.33 & 91.80 & 94.54 & 96.32 \\
\midrule
\textbf{ALUM}                    & \textbf{89.90} & \textbf{92.58} & \textbf{94.74} & \textbf{96.46} \\
\bottomrule
\end{tabular}
}
\vspace{-5pt}
\caption{\label{tab:rejection}Experiment results of accuracy-rejection trade-off on RAF-DB. ($\dagger$:  reproduced from the released code.)}
\end{center}
\vspace{-20pt}
\end{table}

\subsection{Multi-Domain Image Classification}
\label{sec:exp-multidomain}

\noindent
\textbf{Datasets}: To verify the effectiveness of our ALUM on tackling \textbf{unknown data distribution shifting}, we evaluate ALUM on two widely-used multi-domain generalization benchmarks, PACS~\cite{li2017deeper} and Office-Home~\cite{venkateswara2017deep}. For PACS, it contains 4 different style domains (\emph{Art Painting}, \emph{Cartoon}, \emph{Photo}, and \emph{Sketch}), and 7 common categories (dog, elephant, giraffe, guitar, horse, house, and person) with a total number of 9991 images. For Office-Home, it contains 15500 images of 65 classes collected from 4 different domains (\emph{Art}, \emph{Clipart}, \emph{Product}, and \emph{Real}).

\begin{table*}[t]
\begin{center}
\vspace{-10pt}
{\small
\begin{tabular}{l l ccccc ccccc}
\toprule 
\multirow{2}*{Method} & \multirow{2}*{Reference} & \multicolumn{5}{c}{PACS} & \multicolumn{5}{c}{Office-Home} \\ \cmidrule(rl){3-7} \cmidrule(rl){8-12}
&  & \emph{Art} & \emph{Cartoon} & \emph{Photo} & \emph{Sketch} & Ave(\%) & \emph{Art} & \emph{Clipart} & \emph{Product} & \emph{Real} & Ave(\%)\\
\midrule
Mixup~\cite{zhang2018mixup} & ICLR’2018         & 76.8 & 74.9 & 95.8 & 66.6 & 78.5 & 58.2 & 49.3 & 74.7 & 76.1 & 64.6 \\
CrossGrad\cite{shankar2018generalizing} & ICLR'2018 & --- & --- & --- & --- & --- & 58.4 & 49.4 & 73.9 & 75.8 & 64.4 \\
Manifold~\cite{verma2019manifold} &  ICML’2019  & 75.6 & 70.1 & 93.5 & 65.4 & 76.2 & 56.2 & 46.3 & 73.6 & 75.2 & 62.8 \\
CutMix~\cite{yun2019cutmix}  &   ICCV’2019      & 74.6 & 71.8 & 95.6 & 65.3 & 76.8 & 57.9 & 48.3 & 74.5 & 75.6 & 64.1 \\
L2A-OT~\cite{zhou2020learning} & ECCV’2020      & 83.3 & 78.2 & 96.2 & 73.6 & 82.8 & \textbf{60.6} & 50.1 & 74.8 & \textbf{77.0} & 65.6 \\
SagNet~\cite{nam2021reducing}  & CVPR’2021      & 83.6 & 77.7 & 95.5 & 76.3 & 83.3 & 60.2 & 45.4 & 70.4 &  73.4 & 62.3\\
pAdaIN~\cite{nuriel2021permuted} & CVPR’2021    & 81.7 & 76.6 & \textbf{96.3} & 75.1 & 82.5 & --- & --- & --- & --- & ---\\
MixStyle~\cite{zhou2021mixstyle} & ICLR’2021    & 82.3 & 79.0 & 96.1 & 73.8 & 82.8 &  58.7 & 53.4 & 74.2 & 75.9 & 65.5 \\
DSU~\cite{li2022uncertainty}   &  ICLR’2022     & 83.6 & \textbf{79.6} & 95.8 & 77.6 & 84.1 & 60.2 & 54.8 & 74.1 & 75.1 & 66.1 \\
\midrule
\textbf{ALUM}     & \textbf{Ours}               & \textbf{84.3} & 79.4 & \textbf{96.3} & \textbf{81.9} & \textbf{85.5}  &  60.3 & \textbf{54.9} & \textbf{75.3} & 76.6 & \textbf{66.8}\\
\bottomrule
\end{tabular}
\vspace{-5pt}
\caption{\label{tab:pacs&office}Experiment results on the multi-domain image classification dataset PACS and Office-Home. }}
\end{center}
\vspace{-20pt}
\end{table*}

\noindent
\textbf{Settings}: Following the experimental protocols in \cite{zhou2021mixstyle,li2022uncertainty}, we use the same \textit{leave-one-out} protocol (train on three domains and test on the rest one) and ResNet-18 as backbone for all methods. Several SOTA multi-domain classification methods (listed in Table~\ref{tab:pacs&office}) are compared in the experiments, where we show their results reported in the corresponding paper. To train our ALUM, the same training settings and parameters in Sec.~\ref{sec:exp-fer} are used.

\noindent
\textbf{Comparison Results with State-of-the-art Methods}: As shown in Table~\ref{tab:pacs&office}, for PACS, our ALUM outperforms the other multi-domain classification approaches by achieving relatively more stable performance in all four domains (three top 1s, and one top 2s but only slightly worse than the best) so that the overall improvement against SOTA methods is more than 1.4\%. Notably, we significantly improve over SOTA methods on \emph{Sketch} domain with 4.3\%, since \emph{Sketch} is particularly distinct from the other three domains which is challenging for previous methods while the uncertainty modeling of our ALUM can better handle this. Similar results can be seen for Office-Home, with two top 1s, two slightly worse top 2s and overall 0.7\% improvement over SOTA methods. These results demonstrate the utilization of adversarial data uncertainty and latent model uncertainty compensation can better handle unknown domain shifting by obtaining more discriminative and robust domain-invariant features.

\subsection{More Experiments on Image Classification}
\label{sec:exp-general}

\noindent
\textbf{Datasets:}
To verify that the effectiveness of our proposed method on handling noisy labels is not specific for FER, we conduct experiments on common natural image classification datasets CIFAR-10 and CIFAR-100~\cite{krizhevsky2009learning}. To introduce noisy labels, we use CIFAR10N and CIFAR100N~\cite{wei2022learning} with human-annotated real-world noisy labels. CIFAR10 includes six different noisy types (\emph{Clean}, \emph{Aggregate}, \emph{Random 1-3}, \emph{Worst}) while CIFAR100 includes two noisy types (\emph{Clean} and \emph{Noisy}), where \emph{Clean} refers to the original dataset. We follow the training settings in~\cite{wei2022learning} and use ResNet-18 as the backbone for all methods.

\noindent
\textbf{Comparison with Baseline Methods} In Table~\ref{tab:cifar}, we compare our ALUM with the baseline ResNet-18, DSU~\cite{li2022uncertainty} which focuses on modeling latent model uncertainty, and RUL~\cite{zhang2021relative} which focuses on modeling data uncertainty but from a pairwise manner. Since our ALUM explores the interdependency of data uncertainty and model uncertainty as a whole and the proposed adversarial triplet effectively utilizes the uncertainty in data learning, we significantly improve over the Baseline without uncertainty modeling and two closest baseline methods of our ALUM.

\begin{table*}[t]
\begin{center}
{\small
\begin{tabular}{l|cccccc|cc}
\toprule
Method & \multicolumn{6}{c|}{CIFAR-10N} & \multicolumn{2}{c}{CIFAR-100N} \\
\midrule
Noisy Type & \emph{Clean} & \emph{Aggregate} & \emph{Random 1} & \emph{Random 2} & \emph{Random 3} & \emph{Worst} & \emph{Clean} & \emph{Noisy} \\
\midrule
Baseline~\cite{he2016deep} & 91.83 & 85.92 & 80.72 & 81.35 & 81.01 & 68.23 & 72.71 & 50.02 \\
DSU~\cite{li2022uncertainty}  & 92.52 & 87.53 & 82.11 & 83.26 & 82.83 & 68.02 & 72.71 & 49.23 \\
RUL~\cite{zhang2021relative} & 91.71 & 88.16 & 84.75 &  85.20 & 86.15 & 74.49 & 71.05 & 51.73 \\
\midrule
\textbf{ALUM}  & \textbf{93.43} & \textbf{91.04} & \textbf{89.77} & \textbf{90.04} & \textbf{89.22} & \textbf{83.43} & \textbf{74.53} & \textbf{62.29}\\
\bottomrule
\end{tabular}
}
\vspace{-5pt}
\caption{Experiment results on natural image classification dataset CIFAR-10N/100N with different noisy types.}
\label{tab:cifar}
\end{center}
\end{table*}

\subsection{Ablation Study}
\label{sec:exp-ablation}
Now we carry out more ablation analysis to further investigate our proposed ALUM method. 

\noindent
\textbf{The influence of different components}: We conduct experiments to investigate the influence of different components in ALUM. As shown in Table~\ref{tab:ablation}, the pairwise uncertainty modeling method RUL~\cite{zhang2021relative} is compared as Baseline. Four main components (Latent model uncertainty Compensation (LC), Adversarial Positive batch (AP), Adversarial Negative batch (AN), triplet loss $\mathcal{L}_T$ (Triplet) in ALUM are tested, denoted as \textbf{ALUM-1}-\textbf{ALUM-5}. All different versions of ALUM method can improve the test accuracy against the Baseline, and the full model achieves the best performance. These results show the effectiveness of our proposed adversarial data uncertainty modeling and latent model uncertainty compensation modules.

\begin{table}[t]
\begin{center}
\vspace{-10pt}
{\small
\begin{tabular}{l|cccc|c}
\toprule
Model & LC & AP & AN & Triplet & Accuracy(\%) \\
\midrule
Baseline    &  &  &  &  & 80.87 \\
ALUM-1      & \checkmark &  &  &  & 83.12  \\ 
ALUM-2      & \checkmark & \checkmark &  &  &   83.77       \\ 
ALUM-3      & \checkmark &  & \checkmark  &  &   83.80       \\ 
ALUM-4      & \checkmark & \checkmark & \checkmark &  &         83.96 \\
\textbf{ALUM-5}      & \checkmark & \checkmark & \checkmark & \checkmark  & \textbf{84.91} \\ 
\bottomrule
\end{tabular}
}
\vspace{-5pt}
\caption{\label{tab:ablation}The influence of each component in ALUM on RAF-DB with noisy level 30\%.}
\end{center}
\vspace{-10pt}
\end{table}


    

    
\begin{figure}[t]
\vspace{-10pt}
\centering
\includegraphics[width=\linewidth]{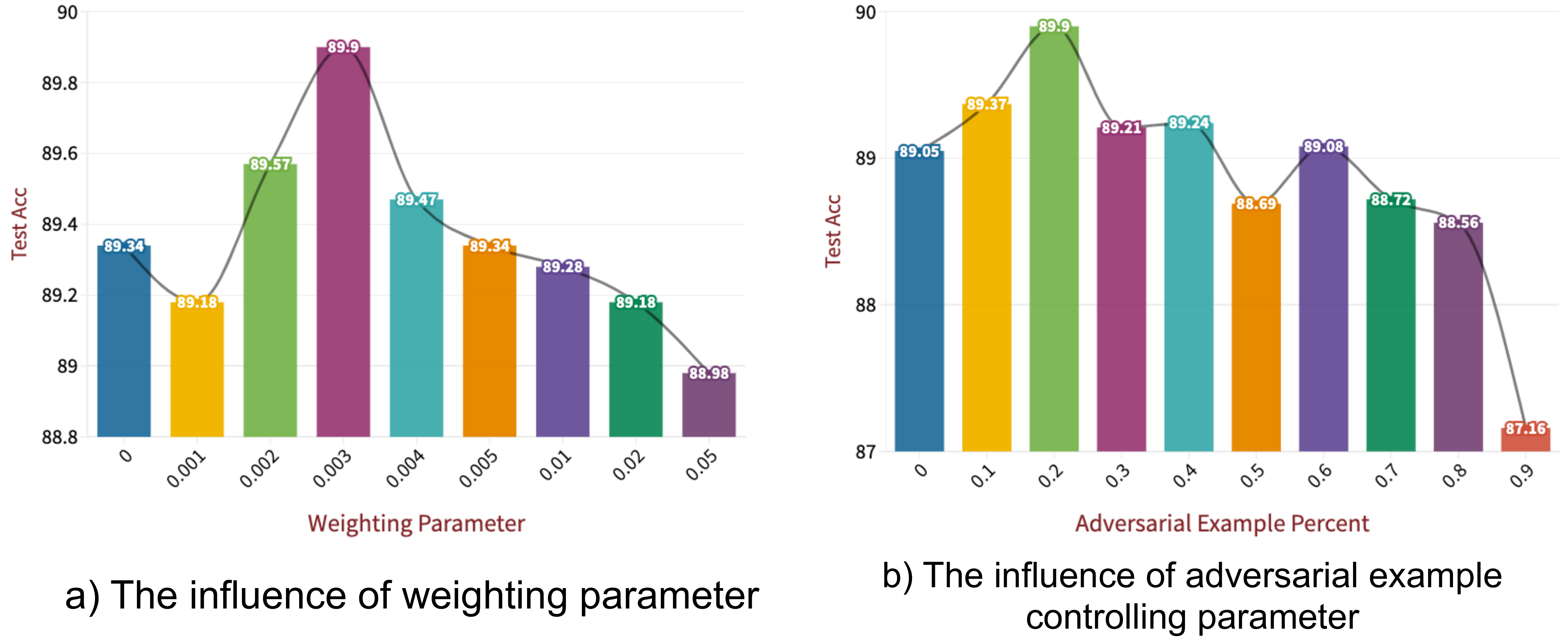}
\vspace{-20pt}
\caption{Ablation studies on the influence of the weighting parameter $\lambda$ and adversarial example percent $p$ on RAF-DB with noisy level 0\%.}
\label{fig:ablation}
\vspace{-10pt}
\end{figure}

\noindent
\textbf{The influence of weighting parameter $\lambda$}:
As indicated by Fig.~\ref{fig:ablation} left, the parameter $\lambda$ that balances the two loss terms in our final objective (Eqn.~\ref{eqn:loss}) has a trade-off effect. A relatively small $\lambda$ cannot fully explore the function of adversarial examples while a relatively large $\lambda$ will over-emphasize the separability of adversarial triplets, but weaken the discrimination ability of the learned model. Therefore, we set $\lambda$ as 0.003 to achieve the best performance. 

\noindent
\textbf{The influence of adversarial example controlling parameter $p$}:
As demonstrated by Sec.~\ref{sec:final}, although the mined adversarial batches $\mathcal{B}_+$ and $\mathcal{B}_{-}$ are informative enough to mitigate the uncertainty diminishing problem, they could also deteriorate the convergence of model training since they are too hard to handle. Therefore, inspired by the self-paced learning theory~\cite{kumar2010self}, we adopt the adversarial examples $\mathcal{B}_+$ and $\mathcal{B}_{-}$ in a ``from-easy-to-hard'' manner. A controlling parameter $p$ is used to control the percentage of how many adversarial samples are mined, and the other samples are the randomly shuffled results of $\mathcal{B}$. As shown in Fig.~\ref{fig:ablation} right, even for a small portion of adversarial examples ($p\le0.4$), the performance of the learned model can be largely improved which demonstrates the effectiveness of adversarial examples to facilitate uncertainty modeling. Keep increasing $p$, more adversarial examples are involved so that the model can not be well-trained based on them. Therefore, a high percentage $p$ of adversarial examples will not benefit uncertainty modeling anymore.

\noindent
\textbf{Computation overhead}: We compare the computation overhead with our closest baseline RUL~\cite{zhang2021relative}, under the same network architecture. On RAF-DB dataset, the average training time for each epoch is 16.68s for RUL and 18.24s for our ALUM. The extra overhead appears in our novelty --- leveraging the mined adversarial examples, and compensating the latent feature maps based on non-parametric uncertainty estimation. The additional time cost is marginal, given the consistent performance improvement in various experiments.

\section{Conclusion}
\label{sec:conclusion}
Unlike existing uncertainty modeling approaches, our proposed ALUM aims to simultaneously quantify the model uncertainty in latent network layers and the data uncertainty in the ultimate feature space. To achieve this, a non-parametric uncertainty estimation is adopted to compensate for the latent feature maps for mitigating the model uncertainty in training. Meanwhile, adversarial examples are directly mined from the given training data to avoid deteriorating the performance of uncertainty learning. Extensive experiments on different noisy learning tasks have shown the superior robustness of ALUM against various data noise conditions than the other state-of-the-art uncertainty modeling methods. 

Currently, there are no mature theoretical justifications to guarantee the effectiveness of ALUM for improving model robustness. In the future, we hope to thoroughly investigate the theoretical justifications behind the proposed ALUM.

{\small
\bibliographystyle{ieee_fullname}
\bibliography{egbib}
}

\end{document}